%% file: main.tex
\documentclass{article} 
\usepackage{iclr2026_conference,times}

\input{math_commands.tex}

\usepackage{hyperref}
\usepackage{url}
\usepackage{graphicx}
\usepackage[utf8]{inputenc} 
\usepackage[T1]{fontenc}    
\usepackage{url}            
\usepackage{booktabs}       
\usepackage{amsfonts}       
\usepackage{nicefrac}       
\usepackage{microtype}      
\usepackage{xcolor}         
\usepackage{amsmath}
\usepackage{amssymb}
\usepackage{amsfonts}
\usepackage{nomencl}
\usepackage{glossaries}
\usepackage{xspace}
\usepackage{multirow}
\usepackage{pifont}
\usepackage{algorithm}
\usepackage{subcaption}
\usepackage{enumitem}
\usepackage{multicol}
\usepackage{caption}
\usepackage{array}
\usepackage{fp}
\usepackage{makecell}
\usepackage{cases}
\usepackage{color}
\usepackage{algpseudocode}
\usepackage{listings}
\usepackage{mathrsfs}
\usepackage{longtable}
\usepackage{colortbl}
\usepackage{bm}
\usepackage{makecell}
\usepackage[table]{xcolor}
\usepackage{wrapfig}
\usepackage[most]{tcolorbox}
\usepackage{subcaption} 
\tcbuselibrary{skins,xparse,breakable}

\newcommand{\ignore}[1]{\iffalse#1\fi}

\newcommand{\ie}{\emph{i.e.,}\xspace}
\newcommand{\eg}{\emph{e.g.,}\xspace}

\newcommand{\paratitle}[1]{\vspace{1.5ex}\noindent\textbf{#1}}

\title{Unleashing Perception-Time Scaling to \\ Multimodal Reasoning Models}

\author{Yifan Li$^{1,2,}$\thanks{Equal contribution} \quad  Zhenghao Chen$^{3,}$\footnotemark[1] \quad  Ziheng Wu$^{3,}$\footnotemark[1]\quad Kun Zhou$^4$ \quad  Ruipu Luo$^{3}$ \\ \textbf{Can Zhang}$^3$ \quad \textbf{Zhentao He}$^3$ \quad \textbf{Yufei Zhan}$^5$ \quad \textbf{Wayne Xin Zhao}$^{1,2,}$\thanks{Corresponding authors} \quad\textbf{Minghui Qiu}$^{3,}$\footnotemark[2] \quad \\
$^{1}$Gaoling School of Artificial Intelligence, Renmin University of China \\
$^{2}$Beijing Key Laboratory of Research on Large Models and Intelligent Governance \\
$^{3}$ByteDance 
$^{4}$University of California, San Diego\\
$^{5}$Institute of Automation, Chinese Academy of Sciences \\
\texttt{\{liyifan0925, batmanfly\}@gmail.com} \\
}

\iclrfinalcopy 
\begin{document}

\maketitle

\begin{abstract}
Recent advances in inference-time scaling, particularly those leveraging reinforcement learning with verifiable rewards, have substantially enhanced the reasoning capabilities of Large Vision-Language Models~(LVLMs). Inspired by this success, similar strategies have been applied to multimodal reasoning, yet their impact on visual perception remains unclear. To investigate this gap, we introduce DisTANCE, a perception-centric benchmark for visual estimation tasks. Evaluation results show that LVLMs exhibit limited estimation precision, and inference-time scaling offers only marginal gains. We attribute this to the fast perception paradigm of current LVLMs, where visual understanding is treated as a one-shot output without modeling the underlying perceptual process. To address this, we propose Perception-Time Scaling (PTS), a novel paradigm that encourages token-rich perception and decomposes complex perception problems into intermediate tractable sub-problems, thereby enabling perception to align with and benefit from inference-time scaling. Combined with reinforcement learning techniques, PTS significantly improves perception accuracy, raising high-precision performance on DisTANCE from 8.0\% to 64.7\%, and generalizes well to out-of-domain tasks. Surprisingly, even though PTS data are purely synthetic, combining them with math reasoning data yields consistent gains in both reasoning and real-world perception benchmarks. Further analysis reveals that PTS introduces more perception-related tokens and increases the model’s attention to image tokens. Our code and data will be publicly released.
\end{abstract}

\input{latex/1_intro}
\input{latex/2_benchmark}

\input{latex/3_method}
\input{latex/4_experiment}

\input{latex/6_conclusion}

\bibliography{iclr2026_conference}
\bibliographystyle{iclr2026_conference}

\newpage
\appendix
\input{latex/7_app}

\end{document}

%% file: math_commands.tex
\usepackage{amsmath,amsfonts,bm}

\def\eqref#1{equation~\ref{#1}}

\def\1{\bm{1}}

\DeclareMathAlphabet{\mathsfit}{\encodingdefault}{\sfdefault}{m}{sl}
\SetMathAlphabet{\mathsfit}{bold}{\encodingdefault}{\sfdefault}{bx}{n}



%% file: latex/1_intro.tex
\section{Introduction}
Recently, reinforcement learning with verifiable rewards (RLVR)~\citep{guo2025deepseekr1} has emerged as a popular paradigm, as it enables models to produce longer chains of thought and thereby improves their reasoning abilities. Such approaches are commonly referred to as inference-time scaling methods~\citep{Zhao-llmsurvey-2023}. Inspired by this success, recent work has extended similar strategies to Large Vision-Language Models (LVLMs), yielding notable gains on challenging multimodal reasoning tasks, such as mathematical and multi-disciplinary domains~\citep{wang2024mathvision, yue2024mmmu}. These approaches encourage models to produce longer and more deliberate reasoning chains. However, their benefits appear largely restricted to the reasoning stage, and it remains unclear whether such gains transfer to perception. Indeed, recent studies suggest that reasoning LVLMs may even be more prone to hallucinations~\citep{liu2025more, yao2025reasoning}.

\begin{figure}[tbp]
    \centering
    \includegraphics[width=\linewidth]{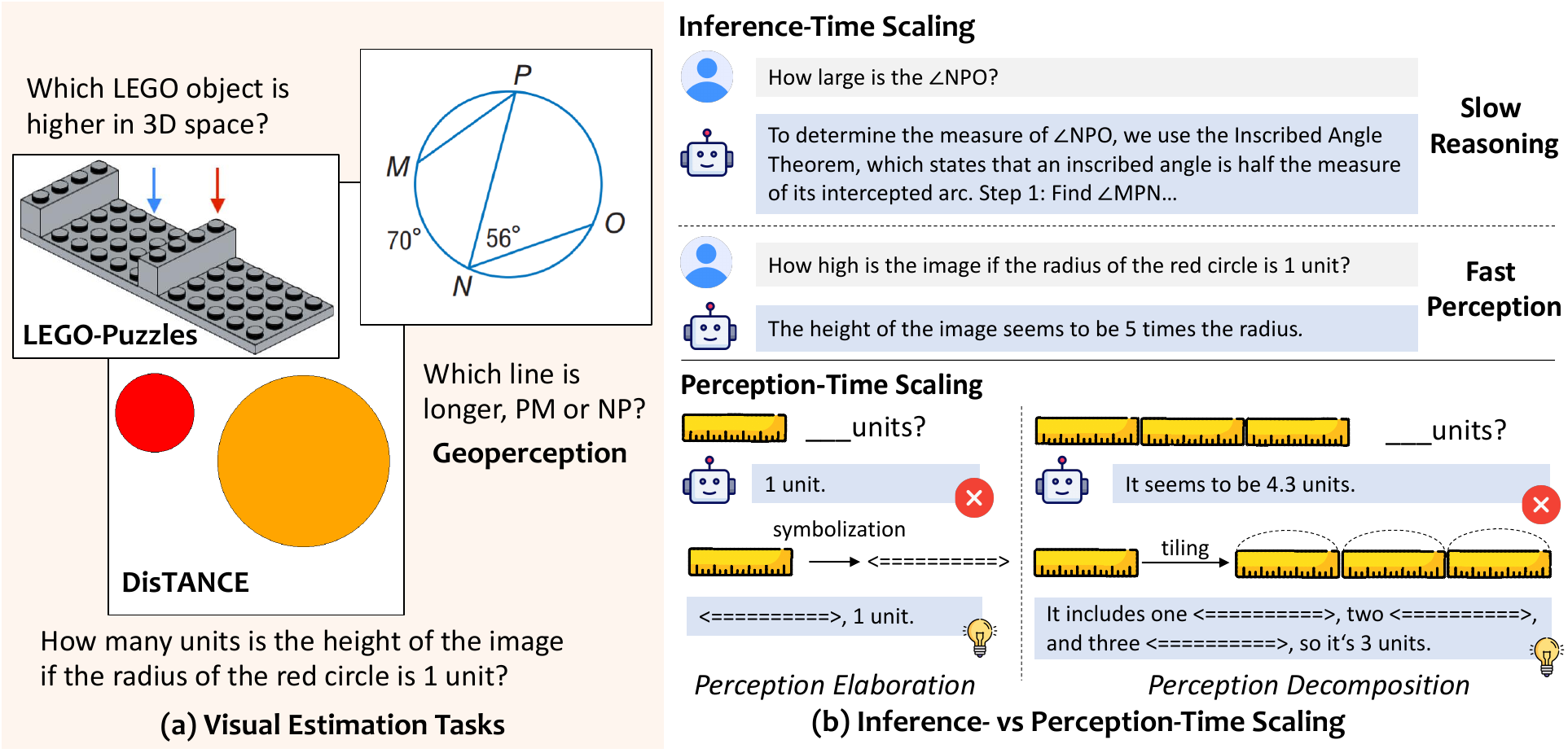}
    \caption{Overview of the visual estimation tasks and comparison between inference-and the proposed perception-time scaling paradigms.}
    \vspace{-10px}
    \label{fig:intro}
\end{figure}

To systematically examine whether perception can also benefit from inference-time scaling, we introduce a perception-centric benchmark, \textbf{DisTANCE} (Distance reasoning Task with Analytical Numeric and Comparative Estimation). DisTANCE comprises synthetic geometric images (\eg circles, triangles) paired with questions that require visual estimation (\eg length, perimeter, area). Evaluation results demonstrate that DisTANCE poses substantial challenges for existing LVLMs, while inference-time scaling methods yield only marginal gains over base models. We attribute this limitation to the prevailing \textit{Fast Perception} paradigm of LVLMs (Figure~\ref{fig:intro}), in which perception is treated as a one-shot judgment rather than a process involving intermediate steps. Further analysis reveals that perception-related tokens constitute only a small fraction of the reasoning chain, and the perception accuracy drops noticeably as task complexity grows.

To tackle this limitation, we propose a new paradigm, \textbf{Perception-Time Scaling} (PTS), which reformulates perception as a structured, step-by-step reasoning process. PTS mitigates the shortcomings of fast perception through two key components: (1) Perception Elaboration, which encourages models to produce token-rich descriptions of perceptual outcomes by using symbolic tokens to represent abstract visual attributes (\eg distance), thereby providing more grounded and interpretable representations than raw numerical values.
(2) Perception Decomposition, which trains models to decompose complex perception tasks into simpler sub-problems.

We instantiate PTS through a two-stage training pipeline. In the first stage, a supervised cold-start phase teaches the model the structured perception pattern. In the second stage, we apply reinforcement learning with GRPO~\citep{shao2024deepseekmath} to further optimize the model, enabling it to explore and refine intermediate perceptual steps. Our experiments demonstrate that PTS enables inference-time scaling to benefit not only reasoning but also perception. Specifically, PTS yields substantial gains on DisTANCE, improving high-precision performance from 8.0\% to 64.7\%, while also generalizing to out-of-domain visual estimation and broader multimodal tasks. Moreover, although PTS data are entirely synthetic, their integration with math reasoning data consistently improves both reasoning and real-world perception benchmarks. Further analysis reveals that PTS introduces more perception-related tokens and increases the model’s attention to image tokens, explaining how it enhances perceptual competence. These findings demonstrate that explicitly modeling perception as a structured process not only enhances interpretability, but also unlocks significant perceptual improvements from inference-time scaling, paving the way toward more perception-aware LVLMs.

%% file: latex/2_benchmark.tex
\section{DisTANCE: Probing Perception of LVLMs with Visual Estimation}

To examine whether inference-time scaling can enhance visual perception, we introduce \textbf{DisTANCE}~(\textbf{Dis}tance reasoning \textbf{T}ask with \textbf{A}nalytical \textbf{N}umeric and \textbf{C}omparative \textbf{E}stimation). 

\begin{table*}[tbp]
  \centering
  \caption{Evaluation results of proprietary, open-source and reasoning-oriented LVLMs on DisTANCE.}
  \scalebox{0.85}{
  \begin{tabular}{lrrrrrrrr}
    \toprule
    \multirow{2.5}{*}{\textbf{Model}} & \multicolumn{2}{c}{\textbf{Length}} &  \multicolumn{2}{c}{\textbf{Perimeter}} & \multicolumn{2}{c}{\textbf{Area}} & \multicolumn{2}{c}{\textbf{Average}} \\
    \cmidrule{2-9}
      & RA$_{0.1}$ & RA$_{\text{avg}}$ & RA$_{0.1}$ & RA$_{\text{avg}}$ & RA$_{0.1}$ & RA$_{\text{avg}}$
      & RA$_{0.1}$ & RA$_{\text{avg}}$ \\
    \midrule
    \rowcolor{gray!20}
    \multicolumn{9}{c}{\emph{Proprietary Models}} \\
    \midrule
    GPT-4o~\citep{openai2024gpt4o}& 22.0& 47.4& 13.0 & 37.6 & \underline{11.0} & \underline{22.0} & 15.3 & 35.7 \\
    Claude-3.5-Sonnet~\citep{claude2024sonnet} & \underline{25.0} & \underline{55.0} & \textbf{23.0} & \underline{55.0} & 9.0 & 15.6& \underline{19.0} & \underline{41.9}\\
    Gemini-2.5-Flash~\citep{google2025gemini} & \textbf{36.0} & \textbf{64.2} & \underline{20.0} & \textbf{57.4} & \textbf{14.0} & \textbf{34.0}& \textbf{23.3}& \textbf{51.9}\\
    Step-R1-V-mini~\citep{stepr1vmini} & 12.0 & 30.2 & 11.0 & 34.4 & 4.0 & 12.6&9.0 & 25.7 \\
    \midrule
    \rowcolor{gray!20}
    \multicolumn{9}{c}{\emph{Open-source Models}} \\
    \midrule
     Qwen2.5-VL-3B~\citep{bai2025qwen25vl} &11.0 & 26.8 & 6.0 & 18.6 & 2.0 & 13.0 & 6.3 & 19.5   \\
    Qwen2.5-VL-7B~\citep{bai2025qwen25vl} & 11.0 & 25.6 & \underline{11.0} & 27.8 & 2.0 & 11.2 & 8.0 & 21.5\\
    Qwen2.5-VL-32B~\citep{bai2025qwen25vl} & 14.0 &38.6& \underline{11.0} & 26.8 & 5.0 & \underline{16.4} & 10.0 & 27.3    \\
    Qwen2.5-VL-72B~\citep{bai2025qwen25vl} & \underline{16.0} & \underline{42.0} & 8.0 & 34.8 & 6.0 & 13.4 & 10.0 & 30.1 \\
    InternVL3-8B~\citep{zhu2025internvl3}    & 11.0 & 29.2 & 8.0 & 21.2 & 5.0 & 11.8 & 8.0 & 20.7\\
     InternVL3-14B~\citep{zhu2025internvl3}     & \underline{16.0} & \underline{42.0}& 10.0 & \underline{37.0} & \underline{9.0} & 14.4 & \underline{11.7} & \textbf{31.1}\\
      InternVL3-38B~\citep{zhu2025internvl3}     & \textbf{19.0} & \textbf{45.2} & 10.0& 30.6& 4.0 & 14.8 & 11.0 & 29.9 \\
      InternVL3-78B~\citep{zhu2025internvl3}     & 11.0 & 35.6 & \textbf{17.0} & \textbf{38.8} & \textbf{10.0} & \textbf{17.2} & \textbf{12.7} & \underline{30.5}\\
    \midrule
    \rowcolor{gray!20}
    \multicolumn{9}{c}{\emph{Open-source Reasoning Models}} \\
    \midrule
    R1-OneVision-7B~\citep{yang2025r1onevision}& 9.0 & 23.4 & \textbf{12.0} &26.0 &\textbf{9.0} & 14.0&\textbf{10.0} & 21.1 \\
    Vision-R1-7B~\citep{huang2025visionr1} & 9.0& 27.0 & 8.0 &25.2 & 6.0 & \textbf{16.8} & 7.7 & 22.7  \\
    MM-Eureka-7B~\citep{meng2025mmeureka} & 6.0 & 20.6 & \underline{10.0} & 27.6 & 3.0 & 10.2& 6.3 & 19.5 \\
    VLAA-Thinker-7B~\citep{chen2025vlaathinker} & \underline{10.0} & \underline{31.8} & 9.0 & \textbf{31.8} & \underline{7.0} & 15.0& 8.7 & \underline{26.2} \\
    NoisyRollout-7B~\citep{liu2025noisyrollout} & \textbf{12.0} &\textbf{34.8} & 9.0 & \underline{30.4} & 6.0 & \underline{15.8}& \underline{9.0} & \textbf{27.0}\\
    \bottomrule
  \end{tabular}}
  \label{tab:distance}
\end{table*}

\subsection{Overview}
DisTANCE focuses on a conceptually simple yet diagnostically effective task: \emph{visual estimation}, which requires models to estimate geometric attributes such as length, perimeter, and area in synthetic images. Since answers rely solely on visual perception, DisTANCE serves as a focused benchmark to test whether current reasoning paradigms also enhance perception.

\paratitle{Task Formulation.} As illustrated in Figure~\ref{fig:intro}, DisTANCE contains tasks that require precise estimation of geometric measurements in synthetic images. Each image includes several geometric objects (\eg circles, triangles, rectangles) with varied colors, sizes, and orientations. The model is prompted to predict quantitative relations between spatial properties in a regression manner (\eg ``What is the height of the image if the radius of the red circle is 1 unit?''), requiring it to perceive and compare distances with numerical precision. To remove confounding effects from image resolution, all tasks are formulated as relative estimations.

\paratitle{Benchmark Details.}
DisTANCE consists of 300 image-question pairs, with each sub-task including 100 examples. All images are automatically synthesized using Python with dimensions ranging from 600 to 1200 pixels, and each image contains 3 to 7 colored shapes. The shapes include circles, triangles, and rectangles filled with distinct colors. Each sub-task has multiple different question templates. More cases are presented in the Appendix \ref{app:qualitative}.

\subsection{Evaluation Setup}
\paratitle{Metrics.}
Following existing work that evaluates LVLMs using numerical answers~\citep{yang2024vsi}, we adopt Relative Accuracy (RA) as the evaluation metric. A prediction $\hat{y}$ is considered correct under threshold $\theta$ if the relative error $|\hat{y} - y| / y < \theta$. We report RA$_\text{avg}$ under various thresholds as:

\begin{equation}
    \text{RA}_{\text{avg}}(\hat{y}) = \frac{1}{|C|}\sum_{\theta \in C} \mathbb{1} \left(\frac{|\hat{y}-y|}{y} < \theta \right),
\end{equation}
where $C = \{0.5, 0.4, 0.3, 0.2, 0.1\}$. We also report RA$_\text{0.1}$ for high-precision performance.

\paratitle{Evaluated Models.}
We evaluate a range of proprietary and open-source LVLMs, including GPT-4o~\citep{openai2024gpt4o}, Gemini-2.5~\citep{google2025gemini}, Claude-3.5~\citep{claude2024sonnet}, Step-R1-v-mini~\citep{stepr1vmini}, Qwen2.5-VL~\citep{bai2025qwen25vl}, and InternVL3~\citep{zhu2025internvl3}, as well as reasoning-oriented LVLMs such as R1-OneVision~\citep{yang2025r1onevision}, Vision-R1~\citep{huang2025visionr1}, MM-Eureka~\citep{meng2025mmeureka}, VLAA-Thinker~\citep{chen2025vlaathinker}, and NoisyRollout~\citep{liu2025noisyrollout}.

\begin{figure}[tbp]
  \centering
  \begin{minipage}[c]{.48\linewidth}
    \centering\small
    \captionof{table}{Comparison on the average response length and perception ratio~(PR) of base and inference-time scaled (ITS) LVLMs.}
    \begin{tabular}{lccc}
      \toprule
      \textbf{Model} & ITS & Avg Len.  & PR \\
      \midrule
      Qwen2.5-VL-7B  & \ding{55} & 199 & 17.6\\
      R1-OneVision-7B& \ding{51} & 322 & 12.4\\
      MM-Eureka-7B  & \ding{51}  & 259 & 13.5\\
      \bottomrule
    \end{tabular}
    \label{tab:perception-ratio}
  \end{minipage}
  \begin{minipage}[c]{.48\linewidth}
    \centering
    \includegraphics[width=\linewidth]{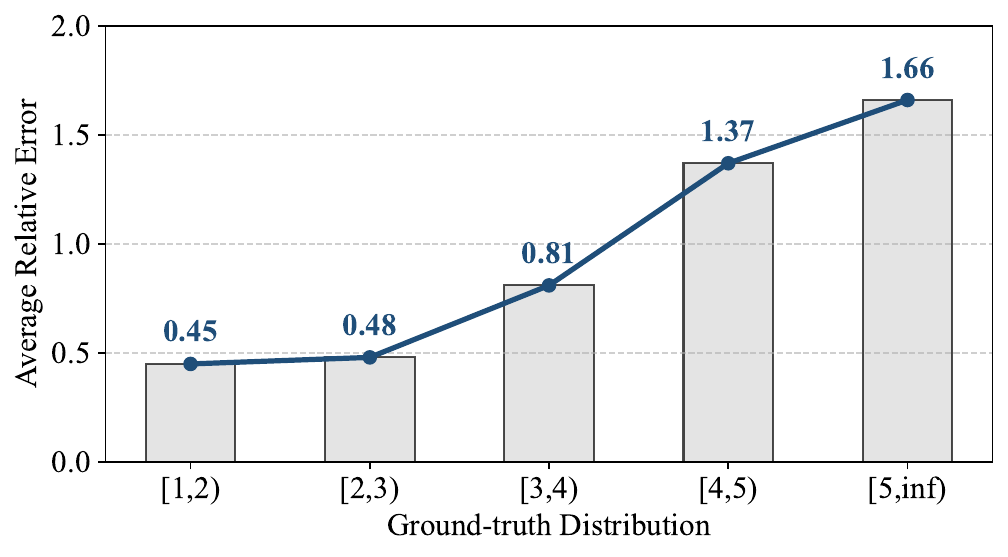}
    \vspace{-20px}
    \captionof{figure}{Relative error trend across labels.}
    \label{fig:error-trend}
  \end{minipage}\hfill
\end{figure}

\subsection{Evaluation Results}
As shown in Table~\ref{tab:distance}, most LVLMs perform poorly on DisTANCE: open-source models rarely exceed an $\text{RA}_{\text{avg}}$ of 35\%, and even frontier systems like Gemini-2.5-Flash reach only 51.9\%. Area estimation is particularly difficult, with most models below 20\%. Scaling model size has limited effect, and reasoning-enhanced LVLMs show only marginal gains over their bases—for example, Vision-R1 (22.7\%) and R1-OneVision (21.1\%) perform on par with Qwen2.5-VL-7B (21.5\%). These results suggest that current inference-time scaling paradigms, although effective at generating longer reasoning chains, do not significantly enhance performance on perception-centric tasks. To further investigate this limitation, we conduct additional analysis and identify two key findings.

\paragraph{LVLMs Prefer Fast Perception.} By observing generated cases, we notice that LVLMs often summarize perceptual information using short, fixed patterns (\eg ``the radius of the circle is 2.5 units''), without further elaboration. We refer to this behavior as \emph{Fast Perception}, a tendency to express perception results in minimal tokens. To quantify this, we analyze the ratio of perception-related tokens (formally defined in Appendix~\ref{app:perception-ratio-definition}) and the average length of model outputs. As shown in Table~\ref{tab:perception-ratio}, reasoning-oriented models produce longer answers but allocate only a small fraction to perception. This highlights a limitation of current inference-time scaling: it scales reasoning but not perception, motivating paradigms that explicitly scale both.


\paragraph{LVLMs Fail to Decompose Complex Perception.}
We further investigate a key failure mode of LVLMs in complex perception tasks. Specifically, we group length-estimation questions by ground-truth distance and measuring Qwen2.5-VL-7B’s relative error. As shown in Figure~\ref{fig:error-trend}, the model’s relative error increases consistently with target length. This trend suggests that while inference-time scaling teaches models to reason step by step, such decomposition ability does not transfer to the perception stage. Even for complex perception cases, models still prefer fast perception, directly producing an answer without any intermediate process or decomposing the task.

%% file: latex/3_method.tex
\section{Methodology}\label{sec:method}
\begin{figure*}[tbp]
    \centering
    \includegraphics[width=\linewidth]{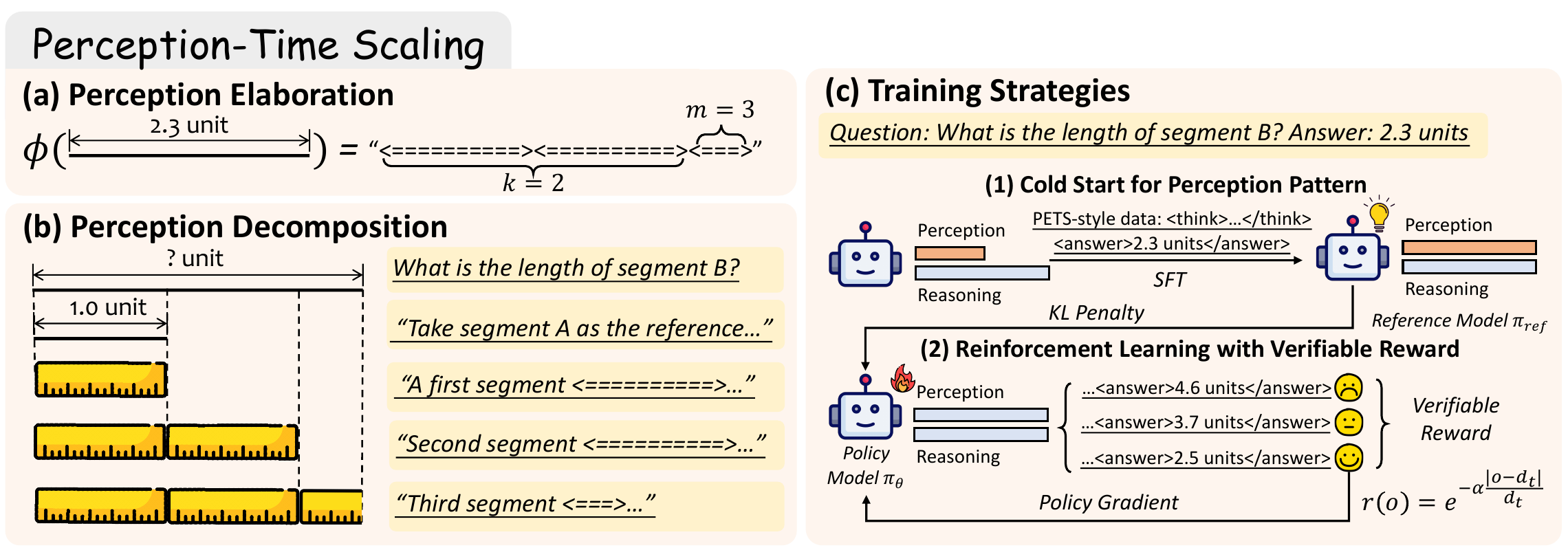}
    \caption{Overview of the proposed PTS paradigm. In the context of visual estimation tasks, PTS first employs symbolic tokens to intuitively represent distance information, followed by a step-by-step decomposition of the estimation process. During training, LVLMs are initially taught the PTS pattern via SFT, then further optimized using GRPO to enhance perception accuracy.}
    \label{fig:pts}
\end{figure*}

Previous empirical results reveal a key limitation: The prevailing fast perception paradigm limits inference-time scaling from improving perception abilities of reasoning-enhanced LVLMs. To address this, we propose a new perception paradigm, \textbf{Perception-Time Scaling (PTS)}, which aims to structurally enhance perception in a way that is compatible with inference-time scaling techniques. In this section, we define the PTS framework and describe how we apply it to the visual estimation tasks from the proposed DisTANCE. The overview of PTS is illustrated in Figure~\ref{fig:pts}. 

\subsection{Perception-Time Scaling}
To overcome the limitations of the Fast Perception paradigm, PTS introduces two complementary strategies. \textbf{(1) Perception Elaboration.} PTS encourages models to produce token-rich descriptions of perceptual outcomes. Specifically, abstract visual attributes such as distance are represented using symbolic tokens, which provide more grounded and interpretable descriptions than raw numerical values. \textbf{(2) Perception Decomposition.} PTS teaches models to decompose complex perception tasks into simpler and more tractable sub-problems. This stepwise approach enables more accurate visual understanding by mimicking how humans often solve challenging perception tasks through progressive comparison. We next describe the detailed implementation of PTS in the context of visual quantitative estimation tasks.

\paragraph{Perception Elaboration.} Given a target distance for estimation $d_t$, we define a symbolic encoding function $\phi$, which maps $d_t$ into a discrete token sequence $\phi(d_t)$. $\phi(d_t)$ consists of several symbolic token groups rendered as \texttt{<==========>}, each representing distance of 1.0 unit. The angle brackets \texttt{<} and \texttt{>} are delimiters and each equal sign \texttt{=} represents a fixed length $\delta=0.1$ unit. Given $d_t\in \mathbb{R}_{\geq 0}$, we decompose it as:
$k = \left\lfloor \frac{d_t}{1.0} \right\rfloor, \quad r = d_t - k,$
where \(k\) is the number of full symbolic token group and $r\in [0, 1.0)$ is the residual length. The encoded sequence is represented as:
\begin{equation}
    \phi(d_t) = \underbrace{\texttt{<==========>}\cdots\texttt{<==========>}}_{k\ \text{times}} \Vert \phi_{\text{res}}(r).
\end{equation}
$\phi_{\text{res}}(r)$ is the symbolization results for $r$, which is defined as:
\begin{equation}
    \phi_{\text{res}}(r) =\texttt{<}\underbrace{\texttt{=}\cdots\texttt{=}}_{m\ \text{times}}\texttt{>}
\end{equation}
where $m = \left\lfloor \frac{r}{\delta} \right\rfloor$. By doing so, PTS increases the presence of perception tokens in model outputs and creates stronger grounding between vision and language. Rather than a single numerical token, the model must now construct the perception output step by step, allowing more intuitive and fine-grained representation of distance information.

\paragraph{Perception Decomposition.} Previous experiments show that even when the task complexity increases, models continue to rely on one-shot perception. To mitigate this limitation, PTS introduces a decomposition strategy that breaks down complex perception tasks into step-by-step local subgoals. Specifically, given the target distance $d_t$, we first select a reference segment $d_r$ from the image, whose length is defined as 1.0 unit. Rather than directly predicting the value of $d_t$, we ask the model to perform a step-by-step accumulation of $d_r$ until fully covering the $d_t$, which can be formulated as:

\begin{equation}
\begin{aligned}
    \text{Initialize:} \quad & L = 0,\quad k = 0; \\
    \text{While } L + d_r \le d_t: \quad & L \leftarrow L + d_r, \quad k \leftarrow k + 1.
\end{aligned}
\end{equation}
This process lets models simulate the process of tiling $d_r$ along $d_t$ and incrementally accumulating distance until the target is fully covered. Each step in this process is relatively simple and localized, making the overall task more tractable. This approach also mirrors how humans handle complex measuring tasks with a ruler. Moreover, by transforming perception into a step-by-step process, this design opens up greater opportunities for inference-time scaling, allowing the model to optimize each perception step and ultimately achieve more accurate visual understanding.

\subsection{Cold Start for Perception Pattern}\label{subsec:data&sft}
We begin by defining the reasoning process under the PTS paradigm. Each reasoning chain follows five stages: (1) \textbf{Review}: A brief description of the task; (2) \textbf{Hint}: Definition for the symbolic encoding and decomposition strategy; (3)\textbf{Reference}: Selection of a specific segment in the image to serve as a reference; (4) \textbf{Estimation}: Visual comparison of other segments to the reference to estimate their lengths; (5) \textbf{Calculation}: Application of appropriate formulas to compute the final result.

To teach LVLMs this structured perception reasoning pattern, we construct a training dataset in PTS style. Following the data collection pipeline of DisTANCE, we synthesize images and corresponding questions for length, perimeter, and area estimation tasks. Each reasoning chain is formatted in line with~\citep{guo2025deepseekr1}, where the reasoning steps are enclosed within \texttt{<think></think>} tags and the final answer within \texttt{<answer></answer>} tags. The dataset is synthesized through a two-step procedure. We first manually collect a small set of reasoning examples for each task, serving as few-shot demonstrations. Next, we feed the distance information within images together with the demonstrations, to GPT-4o to generate instructions. This process yields 2{,}000 examples per task, resulting in a total of 6{,}000 PTS-style reasoning chains.

Importantly, all training images are freshly synthesized with random seeds to avoid overlap with evaluation benchmarks. The resulting dataset is then used for supervised fine-tuning (SFT), enabling the model to acquire an initial understanding of the PTS perception pattern. A full example of the reasoning process and the GPT-4o prompting template is provided in Appendix~\ref{app:data-pipeline}.

\subsection{Reinforcement Learning with Verifiable Reward}\label{subsec:grpo}
To further enhance perception quality under the PTS paradigm, we continue optimizing the model after the cold start stage using GRPO~\citep{shao2024deepseekmath}, a reinforcement learning method with verifiable reward that supports inference-time scaling and improves reasoning capabilities.

\paratitle{Preliminary of GRPO.} GRPO is a reinforcement learning algorithm, which adopts a verifiable reward function instead of a reward model, and calculates the relative advantages within a group of completions for a given sample. Specifically, GRPO first samples a group of answers $\{o_1, o_2, \cdots, o_N\}$ from the old policy model $\pi_{\theta_{\text{old}}}$ for each question $q$. The policy model $\pi_{\theta}$ is then optimized by:

\begin{equation}
\small
    \mathcal{J}(\theta) = \frac{1}{N}\sum_{i=1}^N\left(\min \left(\frac{\pi_{\theta}(o_i|q)}{\pi_{\theta_{\text{old}}}(o_i|q)}A_i, \text{clip}\left( \frac{\pi_{\theta}(o_i|q)}{\pi_{\theta_{\text{old}}}(o_i|q)}, 1-\epsilon, 1+\epsilon\right)A_i\right)-\beta\mathbb{D}_{KL}(\pi_{\theta}||\pi_{\text{ref}})\right),
\end{equation}
where $\epsilon$ and $\beta$ are hyperparameters and $A_i$ is the advantage of $o_i$ in the group. $A_i$ is calculated by:
\begin{equation}
    A_i = \frac{r_i-\text{mean}(\{\ r_1,r_2, \cdots,r_N \})}{\text{std}(\{\ r_1,r_2, \cdots,r_N \})},
\end{equation}
where $r_i$ is the reward given by a rule-based reward function. $r_i$ is typically set as $r = \lambda r_{\text{acc}} + (1-\lambda) r_{\text{format}}$. Here, $r_{\text{acc}}$ is the accuracy reward which is set to 1 if the model's prediction is correct, $r_{\text{format}}$ is the format reward which is set to 1 if the model follows the expected response format, and $\lambda$ is the hyperparameter controls the ratio between two rewards.
In PTS, the perception process is explicitly embedded in the reasoning chain, allowing reward signals to be applied to intermediate perception steps, enabling the model to gradually improve its perception accuracy through step-by-step refinement. We further adapt GRPO to our task with the following modifications:

\paratitle{Customized Reward for Regression.} 
Since tasks in DisTANCE involve continuous regression, binary accuracy rewards are insufficient to capture how close a prediction is to the ground-truth. Moreover, we expect the reward to explicitly encourage high-precision estimation. To address this, we design a continuous reward function based on relative error in exponential form:
\begin{equation}
r(o) = e^{-\alpha \frac{|o-d_t|}{d_t}},
\end{equation}
where $o$ is the model's prediction, $d_t$ is the ground-truth value, and $\alpha$ is a hyperparameter controlling the sensitivity of the reward. This formulation grants higher rewards to predictions with smaller relative errors. The exponential structure makes the reward particularly sensitive to small errors, thus incentivizing fine-grained accuracy. We provide the visualization of the reward function and ablation studies on the choice of $\alpha$ in the Appendix~\ref{app:reward}.

\paratitle{Label Normalization.} We further observe that the same relative error threshold can correspond to very different absolute error tolerances, depending on the scale of the label value. For example, a 10\% relative error threshold on a distance of 0.02 permits an absolute error of only $\pm$0.002, while the same threshold on a value of 50 allows a $\pm$5 deviation.  Consequently, the reward function could assign the same reward to estimations with significantly different levels of precision, which may confuse the model at the beginning of training. To address this, we propose to first train the model on normalized samples with target values less than 1, and later introduce data with random distributions.

%% file: latex/4_experiment.tex
\begin{table*}[tbp]
    \centering
    \caption{Effect of different training data paradigms and strategies on DisTANCE. \emph{Direct}, \emph{CoT}, \emph{PTS} represent the training data in direct answer, Chain-of-Thought and PTS format, respectively. SFT and RL represent whether to adopt supervised fine-tuning or GRPO for training.}
    \scalebox{0.85}{\begin{tabular}{lcccccccccc}
    \toprule
    \multirow{2.5}{*}{\textbf{Model}} & \multirow{2.5}{*}{\textbf{SFT}}&  \multirow{2.5}{*}{\textbf{RL}}&\multicolumn{2}{c}{\textbf{Length}} &  \multicolumn{2}{c}{\textbf{Perimeter}} & \multicolumn{2}{c}{\textbf{Area}} & \multicolumn{2}{c}{\textbf{Average}} \\
     \cmidrule{4-11}
      &&& RA$_{0.1}$ & RA$_{\text{avg}}$ & RA$_{0.1}$ & RA$_{\text{avg}}$ & RA$_{0.1}$ & RA$_{\text{avg}}$
      & RA$_{0.1}$ & RA$_{\text{avg}}$ \\
    \midrule
    SpaceThinker-3B &-&-&7.0 & 39.0 & 8.0 & 38.0 & 6.0 & 26.0 & 7.0 &34.3 \\
    Spatial-R1-7B &-&-&15.0	&51.0& 7.0&	39.0 & 6.0 & 32.0 & 9.3 & 40.7\\
    GPT-4o + DetToolChain &-&-&24.0	&53.8& 16.0&47.8 & 6.0&24.4 & 15.3 & 42.0\\
    GPT-4o + Sketchpad &-&-&27.0	&36.6& 20.0 &	34.5 & 10.0 & 19.2 & 19.0 & 30.1 \\
    \midrule
    Qwen2.5-VL-3B &-&-&11.0 & 26.8 & 6.0 & 18.6 & 2.0 & 13.0 & 6.3 & 19.5   \\
    \midrule
    + \emph{CoT}& \ding{51} & \ding{55}& 15.0 & 48.0 & 28.0 & 61.8 & 13.0 & 34.2& 18.7 & 48.0\\
    + \emph{PTS} &\ding{51} & \ding{55}& 14.0 & 36.0 & 20.0 & 46.8 & 9.0 & 24.6 & 14.3 & 35.8\\
    \midrule
    + \emph{Direct}& \ding{55} & \ding{51}& \textbf{36.0} & \underline{65.6} & 44.0 & 76.8 & 13.0 & 32.8& \underline{35.8} & 47.5\\
    + \emph{CoT}& \ding{51} & \ding{51} & 29.0 &62.7 & \underline{46.0} & \underline{79.2} & \underline{16.0} & \underline{38.7} & 27.0 & \underline{60.2} \\
    \rowcolor{gray!20}
    + \emph{PTS}&\ding{51} & \ding{51} & \underline{31.0} & \textbf{70.6} & \textbf{54.0} & \textbf{86.2} & \textbf{28.0} & \textbf{63.6} &\textbf{37.7} & \textbf{73.5} \\
    \midrule
    Qwen2.5-VL-7B& - & -& 11.0 & 25.6 & 11.0 & 27.8 & 2.0 & 11.2 & 8.0 & 21.5\\
    \midrule
    + \emph{CoT}& \ding{51} & \ding{55}& 17.0 & 40.8 & 10.0 & 52.2 & 14.0 &29.2 & 13.7 & 40.7\\
    + \emph{PTS} &\ding{51} & \ding{55} & 12.0 & 44.2 &25.0 & 62.0 & 12.0 & 30.2 & 16.3 & 35.4\\
    \midrule
    + \emph{Direct}& \ding{55} & \ding{51}& \underline{46.0} &\underline{77.4}&\underline{51.0} & \underline{82.4}&25.0& 52.4& \underline{40.7}&70.7 \\
    + \emph{CoT}& \ding{51} & \ding{51} & 38.0  & 74.6 & 48.0 & 80.8 & \underline{27.0} & \underline{58.6} & 37.7 & \underline{71.3}\\
    \rowcolor{gray!20}
    + \emph{PTS}&\ding{51} & \ding{51} & \textbf{70.0} & \textbf{92.2} & \textbf{74.0} & \textbf{94.6} & \textbf{50.0} & \textbf{78.0} & \textbf{64.7}& \textbf{88.3}\\
    \bottomrule
    \end{tabular}}
    \label{tab:main}
\end{table*}

\section{Experiment}\label{sec:experiment}

\subsection{Experimental Settings}
\paragraph{Baselines.} We compare the proposed PTS against three categories of baselines: 
(1) \textbf{Different data patterns.} Using the same image source, we construct two alternative instruction datasets: one that only includes the final numerical answer (\emph{Direct}), and another that contains conventional chain-of-thought annotations synthesized by GPT-4o (\emph{CoT}). These datasets are used to train the base LVLMs with both SFT and GRPO. 
(2) \textbf{Spatial-aware LVLMs.} Models explicitly designed to estimate spatial information such as object sizes in real-world scenarios, including Spatial-R1~\citep{ouyang2025spatial} and SpaceThinker~\citep{spacethinker}. 
(3) \textbf{Tool-augmented LVLMs.} Approaches that integrate external tools (\eg visual experts or Python code) to provide auxiliary visual information, such as Visual Sketchpad~\citep{hu2024sketch} and DetToolChain~\citep{Wu2024dettoolchain}. Implementation details are presented in Appendix~\ref{app:implementation}.
For backbone models, we select the state-of-the-art LVLMs Qwen2.5-VL-3B and Qwen2.5-VL-7B, and implement our method on them.

\paragraph{Evaluated Benchmarks.}We evaluate model's performance on DisTANCE and other out-of-domain benchmarks: (1) \textbf{DisTANCE$_{\text{ood}}$}: we replace the images in DisTANCE shapes not included in the collected training data (\eg trapezoids and pentagons); (2) \textbf{Geoperception}~\citep{zhang2024euclid}: a fine-grained perception benchmark. We select the LHC subset which requires the model to compare the length of two segments in geometric images; (3) \textbf{LEGO-Puzzles}~\citep{tang2025lego}: a spatial understanding benchmark with LEGO-based tasks, where we select the Height subset that asks the model to distinguish the relative heights of 3D LEGO objects. 

\begin{table*}[tbp]
    \centering
    \caption{Experimental results on out-of-domain distance reasoning tasks.  Geo$_{\text{LHC}}$ is the LineComparison subset from Geoperception. LEGO$_{\text{Height}}$ is the Height subset from LEGO-Puzzles.}
    \begin{tabular}{lcccccccc}
    \toprule
    \multirow{2.5}{*}{\textbf{Model}} & \multirow{2.5}{*}{\textbf{SFT}}& \multirow{2.5}{*}{\textbf{RL}}&\multicolumn{2}{c}{\textbf{Length$_{\text{ood}}$}}&\multicolumn{2}{c}{\textbf{Perimeter$_{\text{ood}}$}} & \textbf{Geo$_{\text{LHC}}$} & \textbf{LEGO$_{\text{Height}}$}  \\
     \cmidrule{4-9}
      &&& RA$_{0.1}$ & RA$_{\text{avg}}$ & RA$_{0.1}$ & RA$_{\text{avg}}$ & Accuracy &Accuracy      \\
    \midrule
    Qwen2.5-VL-3B &-&-& 6.0 & 12.2& 10.0 & 23.8 & 38.9&28.0\\
    \midrule
    + \emph{CoT}& \ding{51} & \ding{55}&8.0 & 29.8 & 16.0 & 45.0 & 64.5& 27.0\\
    + \emph{PTS} &\ding{51} & \ding{55}& 10.0 & 32.6 & 7.0 & 31.8 & 68.4 & 22.0\\
    \midrule
    + \emph{Direct}& \ding{55} & \ding{51}& \underline{12.0} & \underline{38.9}& 23.0 &52.4&70.4&23.0\\
    + \emph{CoT}& \ding{51} & \ding{51} & \underline{12.0} & 37.6 & \underline{24.0} & \underline{53.4}& \underline{71.1}& \underline{30.0}\\
    \rowcolor{gray!20}
    + \emph{PTS}&\ding{51} & \ding{51} & \textbf{16.0} &\textbf{41.4}& \textbf{26.0} & \textbf{55.0} & \textbf{73.5}& \textbf{44.0}\\
    \midrule
    Qwen2.5-VL-7B& - & -& 7.0 & 22.2 & 12.0 &23.6 & 59.8 & \underline{30.0}\\
    \midrule
    + \emph{CoT}& \ding{51} & \ding{55}& 14.0 &33.8 & 21.0 & 52.0& 65.6& 29.0\\
    + \emph{PTS} &\ding{51} & \ding{55} & 15.0 & 36.6 & 26.0 & 56.0 &67.1&28.0 \\
    \midrule
    + \emph{Direct}& \ding{55} & \ding{51}& \underline{19.0} & \underline{40.2} & \underline{30.0} &\underline{63.4} & 69.4&\underline{30.0}\\
    + \emph{CoT}& \ding{51} & \ding{51} & 18.0 & 39.6 & 20.0 & 55.6 &\underline{72.2} &\underline{30.0}\\
    \rowcolor{gray!20}
    + \emph{PTS}&\ding{51} & \ding{51} & \textbf{20.0} & \textbf{43.2} & \textbf{36.0} & \textbf{71.4} & \textbf{78.7}&\textbf{33.0}\\
    \bottomrule
    \end{tabular}
    \label{tab:ood}
\end{table*}

\subsection{Main Results}
We present the experimental results in Table~\ref{tab:main} and ~\ref{tab:ood}, from which we conclude:

\paragraph{PTS Enables High-Precision Visual Estimation.} Across both the 3B and 7B versions of Qwen2.5-VL, models trained with PTS data consistently outperform all baselines. For instance, when combined with SFT and RL, PTS significantly boosts the average $\text{RA}_{\text{avg}}$ of Qwen2.5-VL-7B from 21.5\% to 88.3\%. Notably, under the strict threshold ($\text{RA}_{0.1}$), the model achieves 70.0\%, 74.0\%, and 50.0\% accuracy on the length, perimeter, and area subsets, respectively, demonstrating strong high-precision estimation capabilities. It is also worth noting that PTS strongly outperforms other baselines, despite not using any spatial-related data or external tools. Instead of relying on external modules for spatial perception, PTS internalizes the perception process into the reasoning chain through symbolic tokens, allowing it to be optimized end-to-end during training, which leads to more accurate and robust estimation. Additional qualitative examples are provided in the Appendix~\ref{app:qualitative}.

\begin{table*}[t]
\centering
\caption{Performance comparison of Qwen2.5-VL variants on general multimodal benchmarks.}
\scalebox{0.86}{
\begin{tabular}{lccccccccc}
\toprule
\multirow{2.5}{*}{\textbf{Model}}  &\multirow{2.5}{*}{\textbf{MathVision}}  &\multirow{2.5}{*}{\textbf{MMBench}} &\multirow{2.5}{*}{\textbf{MMVet}}& \multirow{2.5}{*}{\textbf{HalluBench}} &\multirow{2.5}{*}{\textbf{CV-Bench}} & \multicolumn{3}{c}{\textbf{BLINK}} \\
\cmidrule{7-9}
&& &&&&Jigsaw & Multi & Local \\
\midrule
Qwen2.5-VL& 25.3& 83.28 &66.93&51.2& 73.36&66.7 & 46.6 & 48.4\\
+ Geo3K &26.8&83.82& 67.11& 52.2& 74.30 & 66.2& 52.6&52.5\\
+ PTS, Geo3K &\textbf{27.2}  & \textbf{85.68} &\textbf{67.39}&\textbf{53.0}&\textbf{75.68}& \textbf{68.9} &\textbf{53.4} & \textbf{53.3}\\
\bottomrule
\end{tabular}}
\label{tab:general_task}
\end{table*}

\paragraph{PTS Generalizes to Out-of-Domain Estimation Tasks.} 
Table~\ref{tab:ood} demonstrates that PTS exhibits strong generalization capabilities across various estimation tasks beyond the training domain. Specifically, PTS improves the model's estimation performance on geometric shapes not seen during training. It also enhances performance in entirely unseen scenarios. For complex geometric shapes in the Geoperception, PTS boosts the model's accuracy by approximately 20\%. Furthermore, in 3D scenarios from the LEGO-puzzles, PTS improves the model's accuracy in judging the height of objects, highlighting the potential of our method in real-world scenarios.

\begin{figure*}[t] 
    \centering
    \begin{subfigure}{0.45\textwidth}
        \centering
        \includegraphics[width=\linewidth]{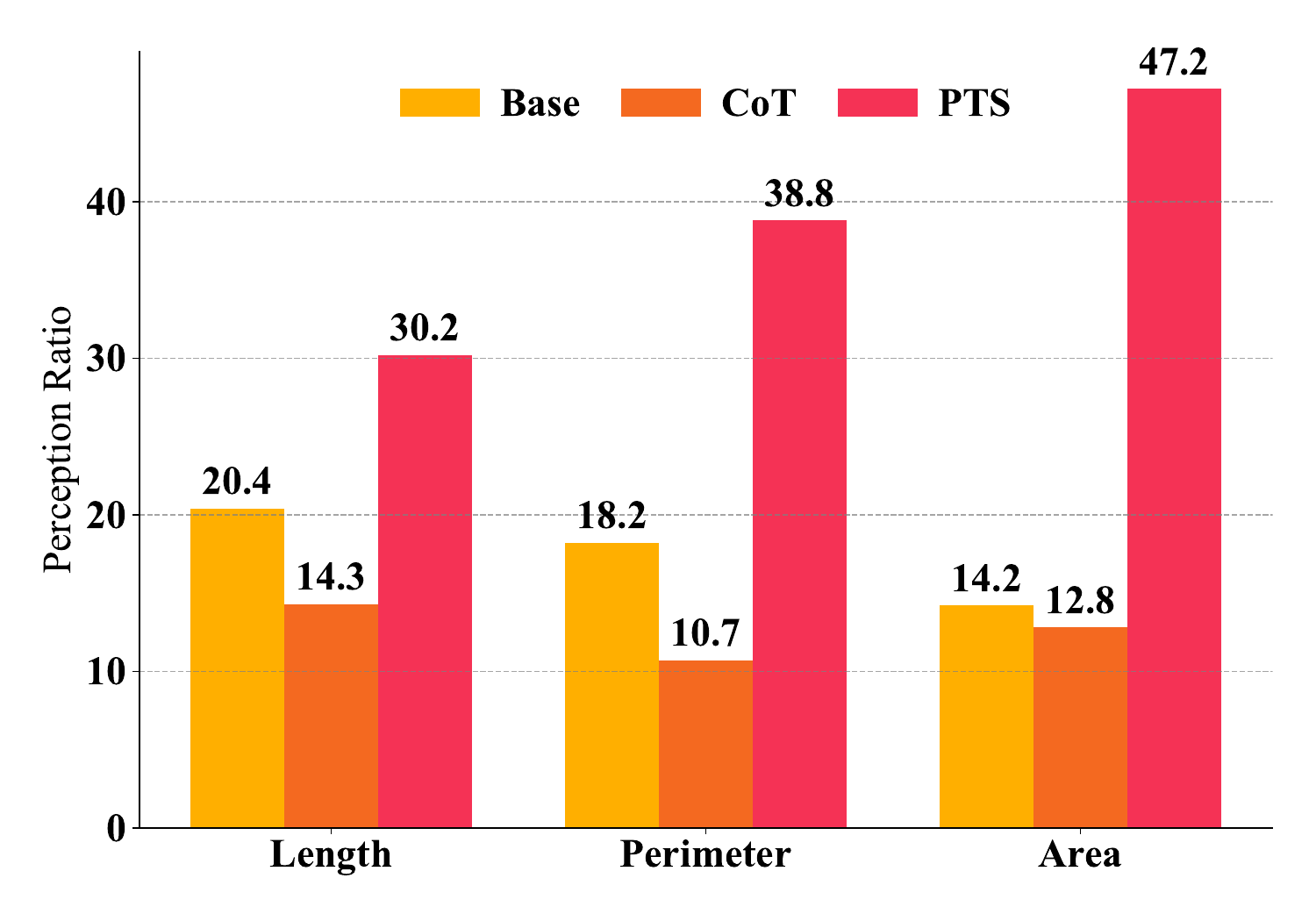}
        \caption{Perception Ratios of Qwen2.5-VL-7B Variants}
        \label{fig:perception_ratio}
    \end{subfigure}
    \hfill
    \begin{subfigure}{0.48\textwidth}
        \centering
        \includegraphics[width=\linewidth]{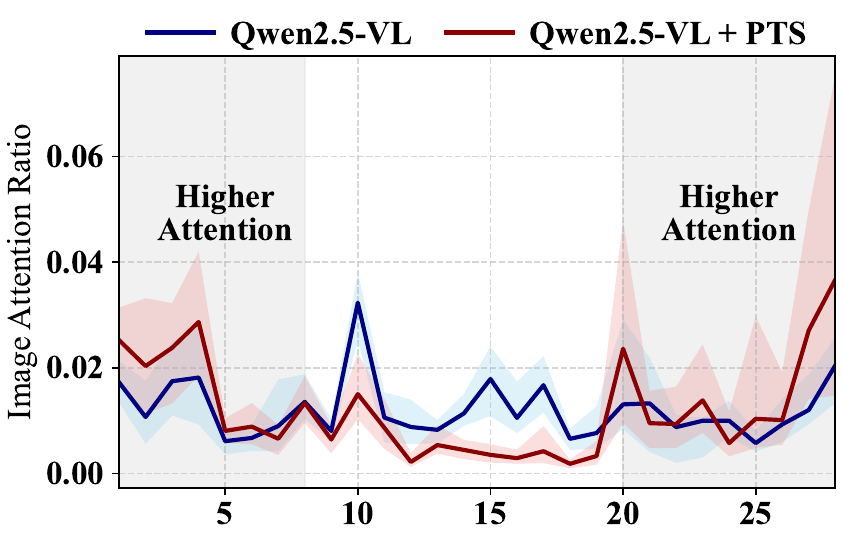}
        \caption{Image attention ratio across transformer layers}
        \label{fig:attention-ratio}
    \end{subfigure}
    \caption{Influence of PTS on model's perception pattern.}
    \vspace{-10px}
\end{figure*}

\subsection{Influence on Perception Capabilities}
To further investigate the influence of PTS on models’ perception capabilities, we validate its effectiveness on more diverse multimodal datasets from multiple domains. We also confirm that PTS data indeed enhances models’ perceptual ability by examining both the explicit perception token ratio and the intrinsic image attention distribution. Ablation studies on symbolic granularity and  key designs and data efficiency analysis are presented in Appendix~\ref{app:ablation} and ~\ref{app:data-efficiency}.

\paragraph{PTS Improves Both General and Perception-Centric Multimodal Tasks.}
PTS further enhances the general multimodal capabilities of LVLMs when combined with math reasoning data from Geometry3K~\citep{lu2021geometry}. We apply GRPO to a cold-started Qwen2.5-VL-7B model and evaluate on MathVision~\citep{wang2024mathvision} for math reasoning, MMBench~\citep{liu2024mmbench} and MMVet~\citep{Yu2024mmvet} for general multimodal VQA, HallusionBench~\citep{guan2024hallusionbench} for hallucination, and CV-Bench~\citep{Tong2024Cambrian} and BLINK~\citep{Fu2024blink} for perception-centric tasks. As shown in Table~\ref{tab:general_task}, models trained only on math reasoning data improve performance on math-related tasks but yield limited or even negative gains on perception-heavy benchmarks. By contrast, incorporating PTS data not only strengthens reasoning ability but also brings consistent improvements across diverse domains—most notably on CV-Bench and BLINK, where fine-grained perception is critical. Remarkably, despite consisting solely of synthetic images, PTS still transfers effectively to real-world perception tasks, highlighting its strong generalization capacity.
\vspace{-5px}
\paragraph{PTS Facilitates Effective Perception Refinement.}While both PTS and CoT yield similar performance after SFT, their differences become evident during reinforcement learning. PTS sequences explicitly embed perception steps into the reasoning chain, offering a richer structure for optimization. During GRPO training, models can explore and refine these intermediate perception results, leading to substantial improvements in estimation accuracy. In contrast, CoT sequences contain minimal perceptual content, limiting the scope for refinement and resulting in marginal gains. As illustrated in Figure~\ref{fig:perception_ratio}, PTS significantly increases the proportion of perception-related tokens, offering RL more opportunities to enhance perceptual accuracy.
\vspace{-5px}
\paragraph{PTS Strengthens Image-Focused Attention.}
We further investigate how PTS training reshapes the model’s attention distribution over image tokens. Specifically, we compare the attention ratio of the first perception token in vanilla Qwen2.5-VL with that of the symbolic tokens introduced by PTS-trained models. As shown in Figure~\ref{fig:attention-ratio}, PTS-trained models consistently assign higher attention to image regions, especially in the early and final layers. This pattern suggests that PTS enhances low-level grounding at the input stage while also reinforcing image-conditioned reasoning in the later decoding process. Together, these results indicate that PTS encourages the model to more effectively integrate visual evidence throughout the perception–reasoning pipeline.

%% file: latex/6_conclusion.tex
\section{Conclusion}
In this paper, we investigated how inference-time scaling techniques affect the perception capabilities of LVLMs. To this end, we introduced DisTANCE, a lightweight yet diagnostic benchmark designed to evaluate LVLMs with visual estimation tasks. Our evaluation revealed that, under the prevailing fast perception paradigm in LVLMs, inference-time scaling offers limited perception improvements. To address this limitation, we proposed PTS, a novel paradigm that makes the perception process more structured and explicit via perception elaboration and decomposition. After two-stage inference-time scaling with SFT and RL, PTS enables substantial perception improvements. Empirical results showed that PTS consistently enhances performance on both in-domain and out-of-domain visual estimation tasks. Further analysis confirmed the generality of the PTS in general multimodal tasks, suggesting a promising direction for future research on inference-time-enhanced perception in LVLMs.

%% file: latex/7_app.tex
\section{Use of Large Language Models}
In this paper, we use large language models only for minor editing tasks in writing, \ie improving readability and grammar, and for occasional debugging hints in coding. The core contributions, including the proposed benchmark, algorithm design, experimental setup, analysis, and conclusions, were fully developed and verified by the authors.

\section{Definition of Perception Ratio}\label{app:perception-ratio-definition}

We define \emph{perception tokens} as those directly describing visual content in the input image, including:
\begin{enumerate}[leftmargin=*]
    \item \textbf{Explicit geometric attributes}, such as shape mentions (\eg ``circle,'' ``line,'' ``triangle'');
    \item \textbf{Numerical values tied to measurements}, such as ``2.5 units,'' ``length of 3 cm,'';
    \item \textbf{Symbolic markers} introduced by our framework (\eg ``\texttt{<====>}'' tokens).
\end{enumerate}
We implement this definition using a keyword-based filter over tokenized model outputs. For each model, we randomly sample 50 generated outputs for calculation.

\begin{figure}[htbp]
  \centering
  \begin{minipage}[c]{.48\linewidth}
  \vspace{0pt}
    \centering
    \includegraphics[width=\linewidth]{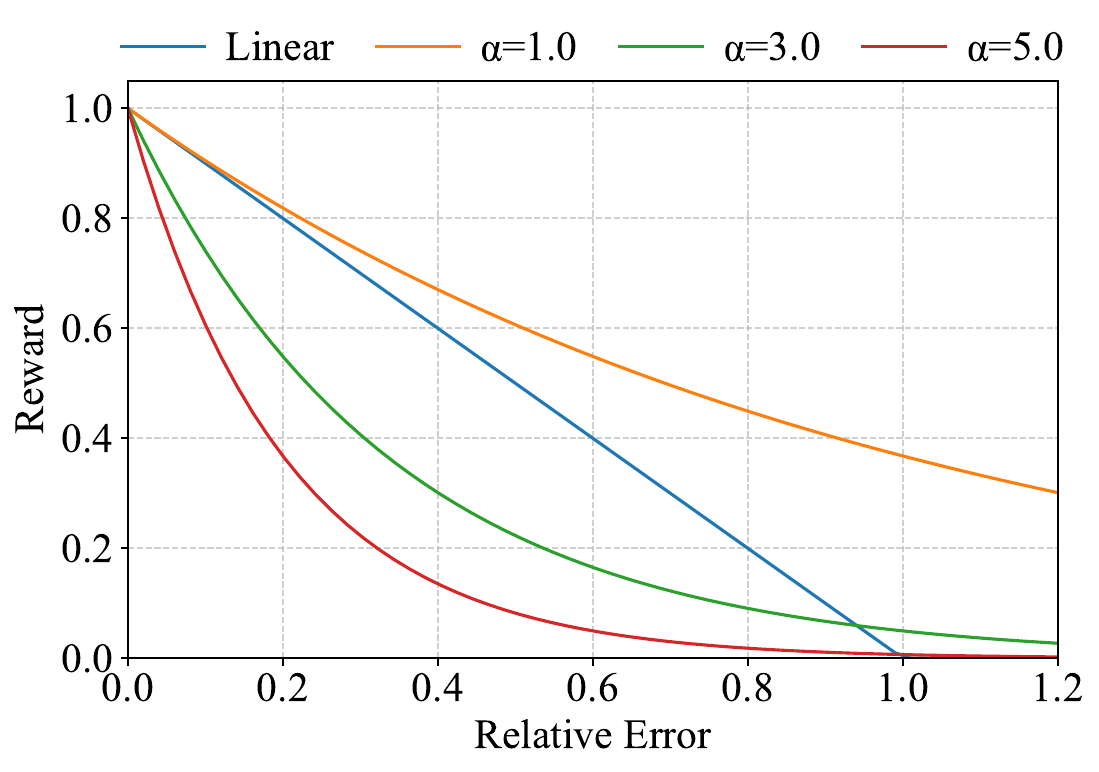}
    \vspace{-10px}
    \captionof{figure}{Curve of linear reward and exponential reward with different $\alpha$.}
    \label{fig:reward}
  \end{minipage}\hfill
  \begin{minipage}[c]{.50\linewidth}
  \vspace{0pt}
    \centering
    \captionof{table}{Performance under varying $\alpha$}
    \begin{tabular}{lcc}
            \toprule
            \multirow{2.5}{*}{\textbf{Parameter}} & \multicolumn{2}{c}{\textbf{DisTANCE}} \\
            \cmidrule{2-3}
            & RA$_{0.1}$ & RA$_{\text{avg}}$ \\
            \midrule
            $\alpha=1.0$ & 53.0 & 75.2\\
            $\alpha=3.0$    & \textbf{64.7} & \textbf{88.3} \\
            $\alpha=5.0$    & 36.0   & 72.4   \\
            \bottomrule
        \end{tabular}
         \label{tab:reward}
  \end{minipage}
\end{figure}

%

\section{Analysis on Reward Factor $\alpha$}\label{app:reward}
We illustrate the curves of the linear reward and the exponential reward under different values of $\alpha$ in Figure~\ref{fig:reward}. As shown, when $\alpha$ is appropriately set (\eg $\alpha=3$ as used in our experiments), the exponential reward remains low and varies smoothly for large relative errors (\eg >0.6), while exhibiting a steep increase for small relative errors (\eg <0.2). Compared to the linear reward, this design encourages the model to make high-precision estimations. We further conduct an ablation study on the choice of $\alpha$ by comparing model performance under $\alpha=1$, $3$, and $5$. As shown in Table~\ref{tab:reward}, both smaller and larger values of $\alpha$ lead to degraded performance compared to our chosen setting. We suggest that when $\alpha$ is too small, the model receives overly high rewards even at the beginning of training, reducing its motivation to improve. Conversely, when $\alpha$ is too large, the reward curve becomes overly flat in the high-error region, where model's predictions commonly fall in the early stages of training, making it difficult for the model to determine a clear direction for improvement.

\begin{figure}[htbp]
  \centering
  \begin{minipage}[c]{.48\linewidth}
  \vspace{0pt}
    \centering
    \captionof{table}{Ablation studies on key designs in PTS.}
    \begin{tabular}{lcc}
            \toprule
            \multirow{2.5}{*}{\textbf{Method}} & \multicolumn{2}{c}{\textbf{DisTANCE}} \\
            \cmidrule{2-3}
            & RA$_{0.1}$ & RA$_{\text{avg}}$ \\
            \midrule
            Qwen2.5-VL-7B-PTS  & \textbf{64.7} & \textbf{88.3} \\
            + symbolic granularity & 62.7 & 83.5 \\
            - exponential reward    & 60.0   & 83.0  \\
            - normalization    & 58.0   & 81.5  \\
            \bottomrule
        \end{tabular}
    \label{tab:ablation}
  \end{minipage}\hfill
  \begin{minipage}[c]{.48\linewidth}
  \vspace{0pt}
    \centering
    \includegraphics[width=\linewidth]{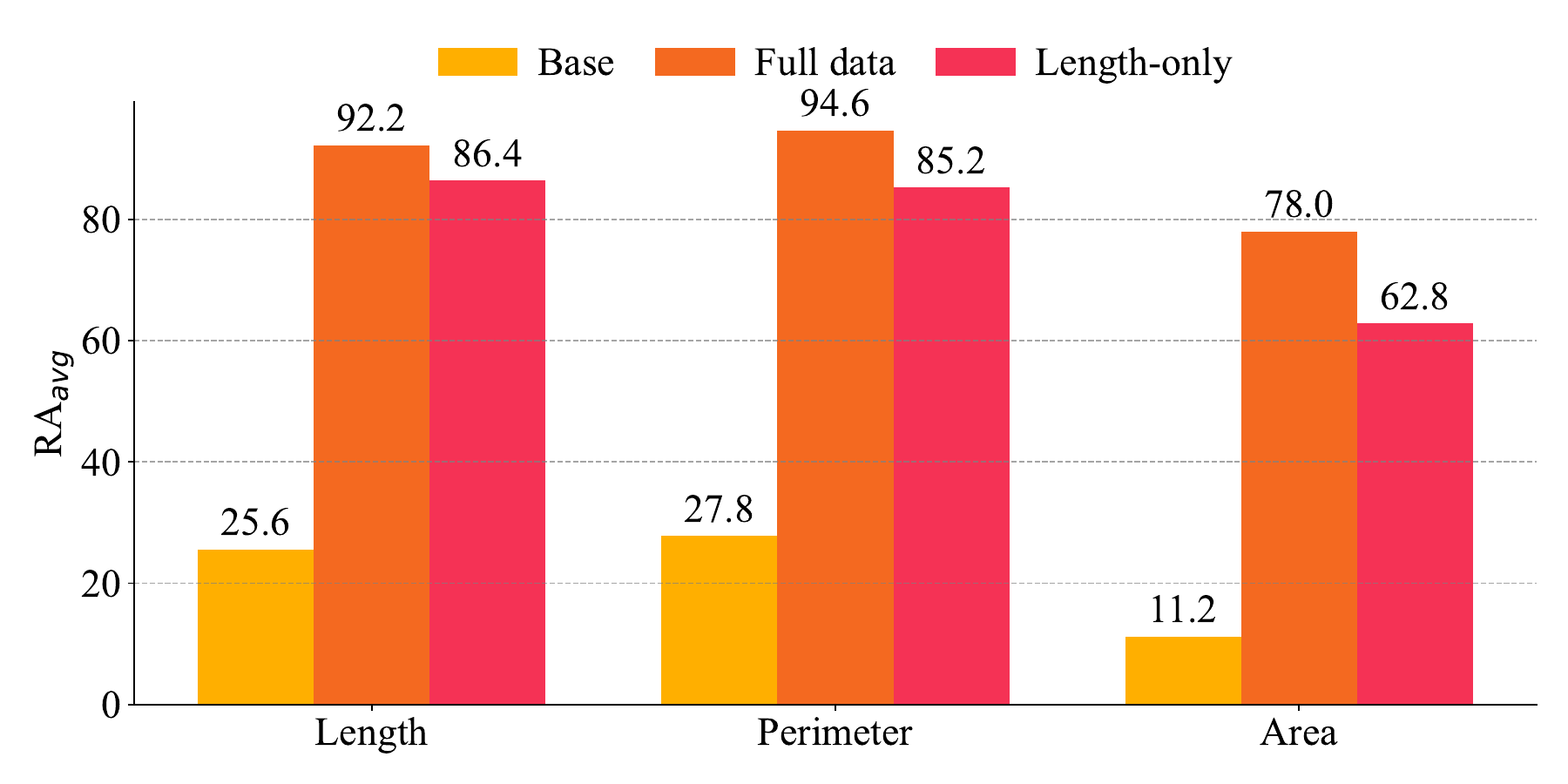}
    \captionof{figure}{Comparison of the performance of models trained with full/length-only PTS data.}
    \label{fig:length}
  \end{minipage}
\end{figure}

\section{Ablation Studies}\label{app:ablation}

We analyze the effectiveness of key components in our method. As shown in Table~\ref{tab:ablation}, removing any of these components negatively affects the model's performance. Replacing the original exponential reward function with a linear one reduces the model’s high-precision accuracy, likely due to its weaker discriminative ability across different relative error levels. Moreover, training without normalizing the ground-truth also results in a performance drop, highlighting the importance of label normalization in stablizing the training process. These results collectively validate the necessity of the designs in PTS. Beyond these ablations, we further experimented with increasing the granularity of symbolic tokens (\eg using “=” to denote 0.05 units instead of 0.1 units). While this finer division also yields competitive results, performance slightly decreases compared to the 10-division strategy. We hypothesize that 10 equal partitions align more closely with common metric conventions, making them more effective for guiding the model’s perception.

\section{Data Efficiency Analysis}\label{app:data-efficiency}
We further investigate the data efficiency of PTS by training models using only length estimation data. As shown in Figure~\ref{fig:length}, despite being restricted to a single sub-task, the model achieves estimation precision on length, perimeter, and area tasks that is highly comparable to the model trained on the full dataset. This finding suggests that equipping LVLMs with a strong ability in primitive distance estimation enables them to generalize naturally to more complex compositional tasks involving distance aggregation (\eg perimeter) or multiplication (\eg area). In particular, length-only training still yields substantial gains over the base model across all sub-tasks, highlighting the central role of length estimation as a foundational skill.

\section{PTS Data Collection Details}\label{app:data-pipeline}

\subsection{Instruction Case}
Below is an instruction example in the PTS format.

\begin{center}
\begin{tcolorbox}[colback=blue!5!white,colframe=blue!55!black,width=0.98\linewidth,breakable,title={\textit{PTS Paradigm}}]
{
    {
        <think>Okay, let me try to figure this out.\\

        <review>\\
        The problem is to estimate the relative length of the orange triangle's longest side compared to the image's height. To solve it, I need to estimate the lengths of both and divide the length of the orange triangle's longest side by the image's height.\\
        </review>\\
        
        <hint>\\
        The task is a precise measurement one. It requires accurately estimating the dimensions within the image. To solve such tasks, I can pick a dimension as the reference and compare other dimensions to it for estimation. To represent length more accurately, I'll use text tokens to visualize the dimensions in the image. The reference line should be 1.0 unit long. I'll represent it as <==========>. The <> are the delimiters and each = is a minimum length unit of 0.1 units long.\\
        </hint>\\
        
        <reference>\\
        In this case, the dimensions I need to estimate are the orange triangle's longest side and the image's height. Let's use the longest side of the orange triangle as the reference. I define its length as 1.0 unit and represent it as <==========>.\\ 
        </reference>\\

        <estimation>\\
        Next, I need to estimate the image's height. I visually compare the image's height to the reference line. The image's height seems to be:\\
        
        First segment <==========> (1.0 unit)\\
        Second segment <=========> (0.9 units)\\
        
        So the height is 1.0 + 0.9 = 1.9 units.\\
        </estimation>\\
        
        <calculation>\\
        Now for the math: the length of the triangle's side (1.0) divided by the image height (1.9). 1.0 / 1.9 = 0.526, which rounds neatly to 0.53.\\
        </calculation>\\

        Yeah, that checks out. So the answer should be 0.53.</think>\\
        
        <answer>0.53</answer>
    }
}
\end{tcolorbox}
\end{center}

\subsection{Prompt for Synthesizing PTS Instructions} \label{app:gpt-prompt}
\begin{itemize}[leftmargin=*]
    \item For \emph{perimeter} estimation, the PTS chain explicitly computes the length of each polygonal edge before summing them.
    \item For \emph{area} estimation, the PTS chain first identifies the appropriate geometric formula (\eg rectangle, triangle, circle) and then estimates the required line segments. 
\end{itemize}

\begin{center}
\begin{tcolorbox}[colback=blue!5!white,colframe=blue!55!black, breakable, width=0.98\linewidth,title={\textit{Prompt for synthesizing PTS instructions}}]
{
    {
        I would like you to convert a json string into a detailed thought process.
        \\

        Your thought process should include the following parts:\\
        1. Review the task (enclosed in <review></review>)\\
        2. Define the estimation methods (select reference line and symbolize dimensions as text tokens) (enclosed in <hint></hint>)\\
        3. Select a side as the reference line (enclosed in <reference></reference>)\\
        4. Visually compare other sides to the reference line and estimate their length (enclosed in <estimation></estimation>)
        5. Use the necessary formulas to calculate the final result (enclosed in <calculation></calculation>)\\
        
        The whole thought process should be formatted as follows: \\
        <think>[Your thought process]</think>\\
        <answer>[Final answer]</answer>\\
        
        Here are some good cases:\\
        input1: \\
        output1: \\

        ...\\
        
        Now, using the following data, please synthesize the thought process. When you mention the information in the json, you MUST pretend that you observe them from an image. Avoid using words like "according to the symbolization result...". No matter how long the line is, you MUST go through every reference segment it contains, instead of using words like "There's a sequence of thirteen <==========> segments".
    }
}
\end{tcolorbox}
\end{center}

\section{Implementation Details}\label{app:implementation}
\subsection{Training Details of PTS}
We select Qwen2.5-VL-3B/7B as the base model. Both models are first supervised fine-tuned with the 6,000 collected PTS data, then further trained by GRPO on another 6,000 answer-only distance reasoning data. We normalize the ground-truth of the first 3,000 data and leave the other data with random distribution. The $\alpha$ is set to 3 in practice. We utilize the implementation of GRPO in EasyR1~\citep{zheng2025easyr1}, with \texttt{rollout\_batch\_size}=512 and \texttt{global\_batch\_size}=128.

\subsection{Baselines}
\begin{itemize}[leftmargin=*]
    \item SpaceThinker~\citep{spacethinker} is trained to enhance spatial reasoning with test-time compute by fine-tuning on synthetic reasoning traces on real-world images with 3D annotations.
    \item Spatial-R1~\citep{ouyang2025spatial}improves LVLMs' spatial reasoning capabilities by introducing the spatial reasoning dataset and task-specific GRPO fine-tuning. 
    \item DetToolChain~\citep{Wu2024dettoolchain} introduces a tool-augmented prompting paradigm that equips LVLMs with zero-shot object detection ability. Since the work has not been open-sourced, we reproduced its Ruler Marker strategy by adding scale marks along the edges of the image and take the \texttt{gpt-4o-0806} as the base model.
    \item Visual Sketchpad~\citep{hu2024sketch} equips LVLMs with a drawing-based reasoning framework, enabling them to sketch lines, boxes, and marks (with optional vision tools) to enhance visual reasoning. We take the \texttt{gpt-4o-0806} as the base model and  set the \texttt{task\_type=vision} for evaluation.
\end{itemize}

\section{Qualitative Analysis}\label{app:qualitative}
\begin{figure}[htbp]
    \centering
    \includegraphics[width=0.5\linewidth]{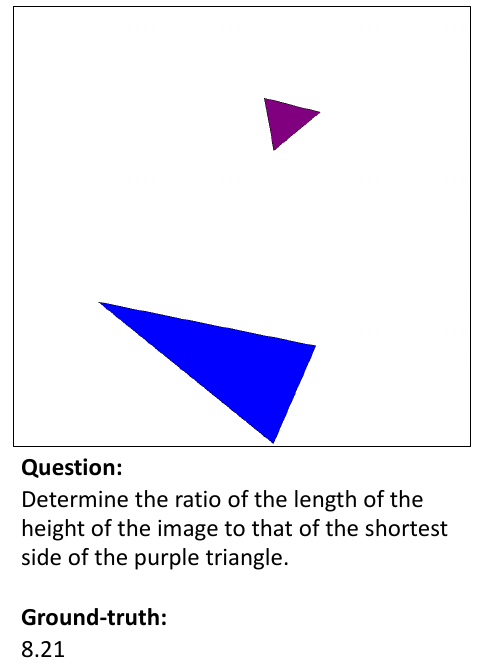}
    \caption{Length estimation case from DisTANCE.}
    \label{fig:case_length}
\end{figure}

\begin{figure}[htbp]
    \centering
    \includegraphics[width=0.5\linewidth]{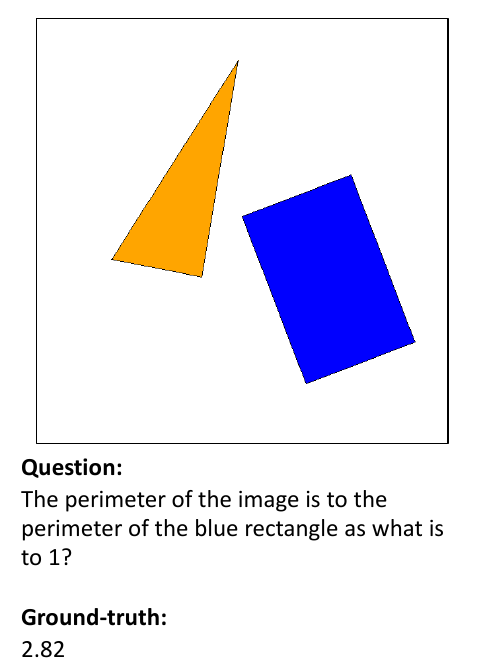}
    \caption{Perimeter estimation case from DisTANCE.}
    \label{fig:case_perimeter}
\end{figure}

\begin{figure}[htbp]
    \centering
    \includegraphics[width=0.5\linewidth]{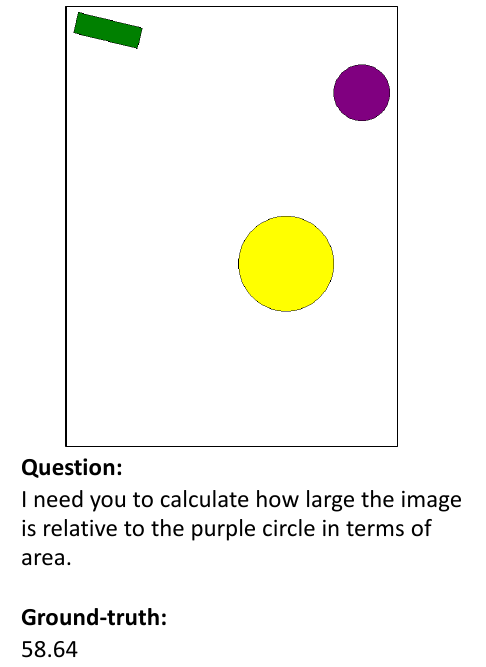}
    \caption{Area estimation case from DisTANCE.}
    \label{fig:case_area}
\end{figure}

In this section, we compare the generation cases of GPT-4o, Qwen2.5-VL-7B and Qwen2.5-VL-7B-PTS on DisTANCE. The questions and images are presented in Figure~\ref{fig:case_length}, \ref{fig:case_perimeter} and \ref{fig:case_area}.

\subsection{Length estimation}

\begin{center}
\begin{tcolorbox}[colback=blue!5!white,colframe=blue!55!black, breakable, width=0.98\linewidth,title={\textit{Length estimation case: Qwen2.5-VL-7B-PTS}}]
{
    {
       <think>Let's tackle this problem step by step.\\
       
       The objective is to determine the ratio of the height of the image to the shortest side of the purple triangle. To achieve this, I need to ascertain the lengths of these two sides and then compute the ratio by dividing the height of the image by the shortest side of the purple triangle.\\
       
       This task involves precise measurement, requiring accurate estimation of the dimensions within the image. A strategic approach is to select one dimension as the reference and compare other dimensions against it for estimation. To visually depict the lengths, I'll use text tokens to represent the dimensions in the image. I'll establish the reference line as 1.0 unit long and represent it as <==========>. Here, the <> are delimiters, and each = symbolizes a minimum length unit of 0.1 units.\\
       
       In this scenario, the two dimensions of interest are the shortest side of the purple triangle and the height of the image. I'll choose the shortest side of the purple triangle as the reference line. Thus, its length is defined as 1.0 unit, represented as <==========>.\\
       
       Now, I need to estimate the height of the image. By visually comparing the image height to our reference line, it can be broken down into segments as follows:\\
       First segment <==========> (1.0 unit),\\
       Second segment <==========> (1.0 unit),\\
       Third segment <==========> (1.0 unit),\\
       Fourth segment <==========> (1.0 unit),\\Fifth segment <==========> (1.0 unit),\\Sixth segment <==========> (1.0 unit),\\Seventh segment <==========> (1.0 unit),\\Eighth segment <==========> (1.0 unit),\\And a final segment <====> (0.4 units).\\Adding these together, the height of the image is 1.0 + 1.0 + 1.0 + 1.0 + 1.0 + 1.0 + 1.0 + 1.0 + 0.4 = 8.4 units.\\For the final calculation, I divide the height of the image (8.4) by the shortest side of the purple triangle (1.0). So, 8.4 $\times$ 1.0 = 8.4.\\Everything checks out. The answer is 8.4.</think>\\\\<answer>8.4</answer>
    }
}
\end{tcolorbox}
\end{center}

\begin{center}
\begin{tcolorbox}[colback=blue!5!white,colframe=blue!55!black,width=0.98\linewidth,breakable, title={\textit{Length estimation case: GPT-4o}}]
{
    {
        To find the ratio of the height of the image to the shortest side of the purple triangle, let's estimate these measurements:\\
        
        1. Height of the Image: The height of the image is the vertical dimension from the top to the bottom. By visually examining the image, this can be approximated as roughly 800 units based on the pixel dimensions.\\
        
        2. Shortest Side of the Purple Triangle: By inspecting the purple triangle, it seems that the shortest side is roughly 30 units in length.\\ 
        
        Now, calculate the ratio:\\
        $\text{Ratio} = \frac{\text{Height of the image}}{\text{Shortest side of the purple triangle}} = \frac{800}{30} \approx 26.67$
        \\
        Therefore, the ratio is \boxed{26.67}.
    }
}
\end{tcolorbox}
\end{center}

\begin{center}
\begin{tcolorbox}[colback=blue!5!white,colframe=blue!55!black,width=0.98\linewidth,breakable, title={\textit{Length estimation case: Qwen2.5-VL-7B}}]
{
    {
       To determine the ratio of the height of the image to the shortest side of the purple triangle, we need to estimate the dimensions of the image and the triangle. Let's assume the image is a rectangle and the triangle is a right triangle with the shortest side being the base.\\
       
       First, let's estimate the dimensions of the image. The image appears to be a rectangle with a height of approximately 10 units and a width of approximately 15 units.\\
       
       Next, let's estimate the dimensions of the purple triangle. The triangle appears to be a right triangle with the shortest side (the base) being approximately 5 units and the height being approximately 8 units.\\
       
       The ratio of the height of the image to the shortest side of the purple triangle is:\\
       
       $\frac{\text{height of the image}}{\text{shortest side of the triangle}} = \frac{10}{5} = 2$ \\
       
       Therefore, the ratio is \boxed{2}.
    }
}
\end{tcolorbox}
\end{center}

\subsection{Perimeter estimation}

\begin{center}
\begin{tcolorbox}[colback=blue!5!white,colframe=blue!55!black,width=0.98\linewidth,breakable, title={\textit{Perimeter estimation case: Qwen2.5-VL-7B-PTS}}]
{
    {
       <think>Let's tackle this problem step by step.\\The task is to determine the relative length of the perimeter of the image compared to the perimeter of the blue rectangle. Essentially, we need to find out how many times the perimeter of the blue rectangle fits into the perimeter of the image. This will involve estimating the perimeters of both shapes and then dividing the perimeter of the image by that of the blue rectangle.\\This task requires precise measurement. It involves accurately estimating the perimeters within the image. To solve such tasks, we can select a dimension as a reference and compare other dimensions to it for estimation. To represent length accurately, we'll use text tokens to visualize the dimensions in the image. The reference line should be 1.0 unit long. We'll represent it as <==========>. The <> are delimiters, and each = is a minimum length unit of 0.1 units long. For the perimeter of the image, we add up the lengths of all its sides (width and height multiplied by 2 since it's a rectangle-like shape). For the blue rectangle, we sum up the lengths of its sides as well.\\In this case, let's take the shorter side of the blue rectangle as our reference. I define its length as 1.0 unit and represent it as <==========>. I'll estimate the lengths of other sides by comparing them to this reference.\\First, I estimate the longer side of the blue rectangle. By visually comparing it to the reference line, it appears to consist of:\\A first segment <==========> (1.0 unit),\\And an additional segment <======> (0.6 units).\\So, the longer side of the blue rectangle is 1.0 + 0.6 = 1.6 units.\\Therefore, the perimeter of the blue rectangle equals 2 * (longer side + shorter side) = 2 * (1.6 + 1.0) = 5.2 units. I estimate the width of the image. By visually comparing it to the reference line, it appears to consist of:\\A first segment <==========> (1.0 unit),\\A second segment <==========> (1.0 unit),\\A third segment <==========> (1.0 unit),\\And a final segment <=======> (0.7 units).\\Adding these together, the width of the image is 1.0 + 1.0 + 1.0 + 0.7 = 3.7 units. I estimate the height of the image. By visually comparing it to the reference line, it appears to consist of:\\A first segment <==========> (1.0 unit),\\A second segment <==========> (1.0 unit),\\A third segment <==========> (1.0 unit),\\A fourth segment <==========> (1.0 unit).\\Adding these together, the height of the image is 1.0 + 1.0 + 1.0 + 1.0 = 4.0 units.\\Therefore, the perimeter of the image equals 2 * (width + height) = 2 * (3.7 + 4.0) = 15.4 units.\\Finally, the relative length of the perimeter of the image compared to the perimeter of the blue rectangle is 15.4 / 5.2 $\approx$ 2.96.\\That seems correct. So the answer should be 2.96.</think>\\\\<answer>2.96</answer>
    }
}
\end{tcolorbox}
\end{center}

\begin{center}
\begin{tcolorbox}[colback=blue!5!white,colframe=blue!55!black,width=0.98\linewidth,breakable, title={\textit{Perimeter estimation case: GPT-4o}}]
{
    {
       To determine the ratio, we need to estimate the perimeters of both the orange triangle and the blue rectangle visible in the image.\\\\1. Estimate the perimeter of the blue rectangle:\\   - It appears as a standard rectangle. Let's estimate dimensions: assume the height is approximately 1.8 times the width.\\   - If width $w = x$, then height $h = 1.8x$.\\   - Perimeter of rectangle $P = 2(w + h) = 2(x + 1.8x) = 3.6x $.\\\\2. Estimate the perimeter of the orange triangle:\\   - Judging by the image, let's assume side lengths roughly proportionate: $a = x $, $b = 1.5x $, $c = 1.2x $.\\   - Perimeter $P = a + b + c = x + 1.5x + 1.2x = 3.7x $, compute the ratio of the total perimeter to just the blue rectangle:$\text{Ratio} = \frac{\text{Perimeter of the orange triangle}}{\text{Perimeter of the blue rectangle}} = \frac{3.7x}{3.6x} = \frac{3.7}{3.6}$
       By calculating the ratio:$\frac{3.7}{3.6} \approx 1.0278$ \\Hence, the answer is:\boxed{1.03}
    }
}
\end{tcolorbox}
\end{center}

\begin{center}
\begin{tcolorbox}[colback=blue!5!white,colframe=blue!55!black,width=0.98\linewidth,breakable, title={\textit{Perimeter estimation case: Qwen2.5-VL-7B}}]
{
    {
       To solve this problem, we need to calculate the perimeters of the orange and blue rectangles and then compare them.\\\\First, let's determine the dimensions of the orange rectangle. The orange rectangle is a 3x4 rectangle, so its perimeter is:\\$ 2 \times (3 + 4) = 2 \times 7 = 14 $, let's determine the dimensions of the blue rectangle. The blue rectangle is a 5x3 rectangle, so its perimeter is:$ 2 \times (5 + 3) = 2 \times 8 = 16 $, we need to find the ratio of the perimeter of the orange rectangle to the perimeter of the blue rectangle. This ratio is:\\$ \frac{14}{16} = \frac{7}{8} = 0.875 $\\\\Therefore, the final answer is:\\ \boxed{0.875}
    }
}
\end{tcolorbox}
\end{center}

\subsection{Area estimation}

\begin{center}
\begin{tcolorbox}[colback=blue!5!white,colframe=blue!55!black,width=0.98\linewidth,breakable, title={\textit{Area estimation case: Qwen2.5-VL-7B-PTS}}]
{
    {
       <think>Alright, let's start working this out.\\The task is to calculate how large the image is relative to the purple circle in terms of area. To achieve this, I need to estimate the areas of both the image and the purple circle, and then divide the area of the image by the area of the purple circle.\\This task demands precise measurement, requiring accurate estimation of areas within the image. I can select a dimension as a reference and compare other dimensions to it for estimation purposes. To represent lengths more precisely, I'll use text tokens to visualize the dimensions in the image. The reference line should be set as 1.0 unit long and represented as <==========>. The <> are delimiters, and each = represents a minimum length unit of 0.1 units. For the area of the image, since it has a rectangle-like shape, we calculate it by multiplying its width and height. For the circle, we use the formula for the area of a circle, which is A = \u03c0r\u00b2, where r is the radius. So, I need to estimate the radius of the purple circle, the width, and the height of the image.\\In this case, let's take the radius of the purple circle as our reference. I'll define its length as 1.0 unit and represent it as <==========>. I will estimate the lengths of other sides by making visual comparisons to this reference line.\\First, I estimate the width of the image. When visually comparing the image width to our reference line, I can break it down as follows:\\There's the first <==========> segment (1.0 unit),\\A second <==========> segment (1.0 unit),\\A third <==========> segment (1.0 unit),\\A fourth <==========> segment (1.0 unit),\\A fifth <==========> segment (1.0 unit),\\A sixth <==========> segment (1.0 unit),\\A seventh <==========> segment (1.0 unit),\\An eighth <==========> segment (1.0 unit),\\A ninth <==========> segment (1.0 unit),\\A tenth <==========> segment (1.0 unit),\\An eleventh <==========> segment (1.0 unit),\\A twelfth <==========> segment (1.0 unit),\\And finally, a <=====> segment, which is 0.5 units long.\\The total width of the image sums up to 1.0 * 12 + 0.5 = 12.5 units. Next, I estimate the height of the image. By visually comparing the image height to our reference line, I can break it down like this:\\There's the first <==========> segment (1.0 unit),\\A second <==========> segment (1.0 unit),\\A third <==========> segment (1.0 unit),\\A fourth <==========> segment (1.0 unit),\\A fifth <==========> segment (1.0 unit),\\A sixth <==========> segment (1.0 unit),...(Omitted due to space limitations.)\\A thirteenth <==========> segment (1.0 unit),\\A fourteenth <==========> segment (1.0 unit),\\A fifteenth <==========> segment (1.0 unit),\\And finally, a <==> segment, which is 0.2 units long.\\The total height of the image sums up to 1.0 * 15 + 0.2 = 15.2 units.\\Consequently, the area of the purple circle is calculated using the formula $A = \pi r^2$, with r = 1.0 and $\pi$ approximately 3.14. So, the area is $3.14 * 1.0^2 = 3.14$ square units.\\For the image, since we know the width is 12.5 units and the height is 15.2 units, using the formula for the area of a rectangle A = width * height, the area of the image is 12.5 * 15.2 = 190.0 square units.\\Finally, the proportion of the image's area relative to the purple circle's area is 190.0 / 3.14 $\approx$ 60.7.\\Yes, that seems right. So the answer should be 60.7.</think>\\<answer>60.7</answer>
    }
}
\end{tcolorbox}
\end{center}

\begin{center}
\begin{tcolorbox}[colback=blue!5!white,colframe=blue!55!black,width=0.98\linewidth,breakable, title={\textit{Area estimation case: GPT-4o}}]
{
    {
       To calculate the area of the image relative to the purple circle, we need to estimate the radius of the purple circle and the dimensions of the image first.\\\\Let's start by estimating the radius of the purple circle. Based on the image:\\- The purple circle appears small in comparison to the image dimensions.\\\\Let's estimate the purple circle's radius as follows:  \\- Assume the radius of the purple circle ($r$) consists of roughly 1/10th of the width of the image.\\- The image width appears to be in a ratio of approximately 4:3 (width:height).\\Now, let's calculate the area of the image and the area of the circle:\\Assume the image width is $800$ pixels and height is $600$ pixels (which are typical dimensions for the aspect observed).\\- Image area = $width \times height = 800 \times 600 = 480,000 \text{ pixels}^2$.\\\\Assuming the estimated radius of the purple circle:\\- $r = \frac{800}{10} = 80 \text{ pixels}$.\\\\Circle area = $\pi r^2 = \pi \times 80^2 = 6,400\pi \text{ pixels}^2$.\\Now calculate the relative area:\\- Relative area = $\frac{Image\\ Area}{Circle\\ Area} = \frac{480,000}{6,400\pi} \approx \frac{480,000}{20,096} \approx 23.89$.\\The relative area of the image compared to the purple circle is approximately:\boxed{23.89}
    }
}
\end{tcolorbox}
\end{center}

\begin{center}
\begin{tcolorbox}[colback=blue!5!white,colframe=blue!55!black,width=0.98\linewidth,breakable, title={\textit{Area estimation case: Qwen2.5-VL}}]
{
    {
       To estimate the size of the image relative to the purple circle in terms of area, we need to compare the areas of the two shapes. However, since the image is not provided, I will use a hypothetical scenario where the purple circle is a standard circle with a radius of 1 unit for simplicity.\\1. Area of the Purple Circle:\\  The area $ A $ of a circle is given by the formula:\\  $\\  A = \pi r^2\\  $\\  For a circle with radius $ r = 1 $:\\  $\\  A = \pi \times 1^2 = \pi \approx 3.14\\  $\\2. Estimating the Area of the Image:\\  The image appears to be a rectangle with a width and height that can be estimated. Let's assume the width and height of the rectangle are both approximately 2 units each (this is a rough estimate based on the visual proportions).\\  The area $ A $ of a rectangle is given by:\\  $\\  A = \text{width} \times \text{height}\\  $\\  For a rectangle with width and height both 2 units:\\  $\\  A = 2 \times 2 = 4\\  $\\3. Comparing the Areas:\\  The area of the image (rectangle) is approximately 4 square units, and the area of the purple circle is approximately 3.14 square units.\\4. Ratio of the Areas:\\  The ratio of the area of the image to the area of the purple circle is:\\  $\\  \text{Ratio} = \frac{\text{Area of the Image}}{\text{Area of the Purple Circle}} = \frac{4}{3.14} \approx 1.27\\  $\\\\Therefore, the image is approximately 1.27 times the area of the purple circle.\boxed{1.27}
    }
}
\end{tcolorbox}
\end{center}

\section{Related Work}\label{sec:related_work}

\paragraph{Multimodal Reasoning Models.} 
Inspired by the success of large reasoning models~\citep{guo2025deepseekr1}, recent research has delved into enhancing the reasoning capabilities of Large Vision-Language Models (LVLMs)~\citep{bai2025qwen25vl, zhu2025internvl3, wu2025valley2}. Early methods~\citep{xu2024llavacot, du2025virgo, yao2024mulberry} focus on collecting multimodal Chain-of-Thought data to conduct supervised fine-tuning on LVLMs. More recently, spurred by the achievements of reinforcement learning algorithms such as GRPO~\citep{shao2024deepseekmath}, some studies have explored the application of reinforcement learning to improve the long-sequence reasoning ability of LVLMs. Some work ~\citep{chen2025r1v, liu2025visualrft, zhan2025visionr1} investigate the design of different verifiable reward functions for various visual tasks. MM-Eureka~\citep{meng2025mmeureka} attempts to directly employ rule-based reinforcement learning to stimulate the model's reasoning and reflective abilities without performing SFT. VLAA-Thinker~\citep{chen2025vlaathinker} explores the influence of cold start to the reasoning capabilities of LVLMs. Our work focuses on adopting RL to enable LVLMs to solve precise measurement problems by including their perception process into text-formed long reasoning chains. 

\paragraph{Visual Perception in LVLMs.} As LVLMs achieve impressive progress in multimodal reasoning tasks, their visual perception accuracy remains a key bottleneck to their real-world applications. Some work studies the hallucination issue by evaluating whether LVLMs accurately describe image contents like object class~\citep{li2023pope} and relationship~\citep{wu2024relationhal}. HallusionBench~\citep{guan2024hallusionbench} further challenges LVLMs with detailed visual illusions, while MMVP~\citep{tong2024eyes} evaluates LVLMs on image pairs that share similar visual representations but significant semantic information. Recently, spatial intelligence~\citep{chen2024spatialvlm} has attracted increasing attention. Existing works~\citep{yang2024vsi, cheng2024spatialrgpt} have proposed benchmarks to evaluate the spatial reasoning abilities of LVLMs, including understanding relative spatial relationships and estimating absolute distances. Unlike real-world estimation benchmarks, DisTANCE consists of 2D synthetic geometric shapes, eliminating the need for commonsense knowledge and concentrating solely on the model's ability to perceive basic distance information.